\documentclass[10pt]{article}
\usepackage[OE]{express}
\usepackage{amsmath,amsfonts,amssymb}
\usepackage{array,multirow,graphicx}
\usepackage{graphicx}
\usepackage{setspace}
\usepackage{tocloft}

\begin{document}

\title{ReLayNet: Retinal Layer and Fluid Segmentation of Macular Optical Coherence Tomography using Fully Convolutional Networks}

\author{Abhijit Guha Roy,\authormark{1,2,3*} Sailesh Conjeti,\authormark{1,*},\\ Sri Phani Krishna Karri\authormark{3}, Debdoot Sheet\authormark{3}, Amin Katouzian\authormark{4}, Christian Wachinger\authormark{2} and Nassir Navab\authormark{1,5}}

\address{\authormark{1}Computer Aided Medical Procedures, Technische Universit\"{a}t M\"{u}nchen, Munich, Germany\\
\authormark{2}Artificial Intelligence in Medical Imaging (AI-Med), Department of Child and Adolescent Psychiatry, Ludwig-Maximilians-University, Munich, Germany\\
\authormark{3}Indian Institute of Technology Kharagpur, WB, India\\
\authormark{4}IBM Almaden Reasearch Center, Almaden, USA\\
\authormark{5}Computer Aided Medical Procedures, Johns Hopkins University, USA}

\email{\authormark{*}abhijit.guha-roy@tum.de, sailesh.conjeti@tum.de, A.Guha Roy and S.Conjeti contributed equally for this work.} %% email address is required



\begin{abstract}
Optical coherence tomography (OCT) is used for non-invasive diagnosis of diabetic macular edema assessing the retinal layers. In this paper, we propose a new fully convolutional deep architecture, termed ReLayNet, for end-to-end segmentation of retinal layers and fluid masses in eye OCT scans. ReLayNet uses a contracting path of convolutional blocks (encoders) to learn a hierarchy of contextual features, followed by an expansive path of convolutional blocks (decoders) for semantic segmentation. ReLayNet is trained to optimize a joint loss function comprising of weighted logistic regression and Dice overlap loss. The framework is validated on a publicly available benchmark dataset with comparisons against five state-of-the-art segmentation methods including two deep learning based approaches to substantiate its effectiveness.
\end{abstract}


%%%%%%%%%%%%%%%%%%%%%%% References %%%%%%%%%%%%%%%%%%%%%%%%%

\section{Introduction}
\label{sec:intro}  % \label{} allows reference to this section
Spectral Domain Optical Coherence Tomography (SD-OCT) is a non-invasive imaging modality commonly used for acquiring high resolution ($6 \mu$m) cross-sectional scans for biological tissues with sufficient depth of penetration ($0.5-2$ mm)~\cite{nassif2004vivo,anger2004ultrahigh}. It uses the principle of speckle formation through coherence sensing of photons backscattered within highly scattering optical media like biological soft tissues~\cite{Huang1991}. It has found its application in medical imaging ranging across retinal pathology investigation, to skin imaging for monitoring wound healing~\cite{yeh2004imaging} and intravascular imaging for effective stent placement~\cite{bouma2003evaluation}, lumen detection~\cite{roy2016lumen} and plaque detection~\cite{roy2016multiscale}. OCT is the preferred modality of choice for cross-sectional imaging of the retina on account of its high resolution favoring clear visualization of the various constituent layers of the retina.

Diabetes is a widely occuring chronic, metabolic disease with an estimated incidence in about $415$ million people (roughly $8.3\%$ of human adult population)~\cite{weng2016decreasing}. Diabetic individuals often are under high risk of developing vision-related co-morbidities, which is reported at a significant $28\%$~\cite{klein1995vision}. The degradation of quality of vision in diabetics is often associated to diabetic retinopathy (DR), which results in damage of retinal blood vessels and accumulation of fluid between the retinal layers~\cite{kumar2014robbins, marmor1999mechanisms}. Thus, proper monitoring of the retinal layer morphology and fluid accumulation is necessary for diabetic patients to prevent chances of occurrence of blindness.

Acquisition of retinal OCT centered at the optical nerve and fovea is highly challenging due to the presence of micro-saccadic eye movements resulting in motion artifacts, variations in tissue inclination with respect to the coherence wave surface and poor signal to noise ratio with increasing imaging depth. The acquisition is also particularly difficult in cases of highly myopic eyes. These inherent challenges associated with the modality makes the interpretation of an OCT image difficult and often highly variable among experts. Specifically, due to the highly diffused nature of the boundaries between two retinal layers. This causes manual annotation of the layer boundaries to be very subjective and time consuming. This has motivated a lot of research for developing automated methods for segmenting different retinal layers from OCT images and aid in accurate diagnosis with minimum subject variation in reporting~\cite{kafieh2013,chiu2010,srinivasan2014,chiu2015kernel,dufour2013graph}. 

\begin{figure}[t]
\centering
\includegraphics[height=0.35\textwidth]{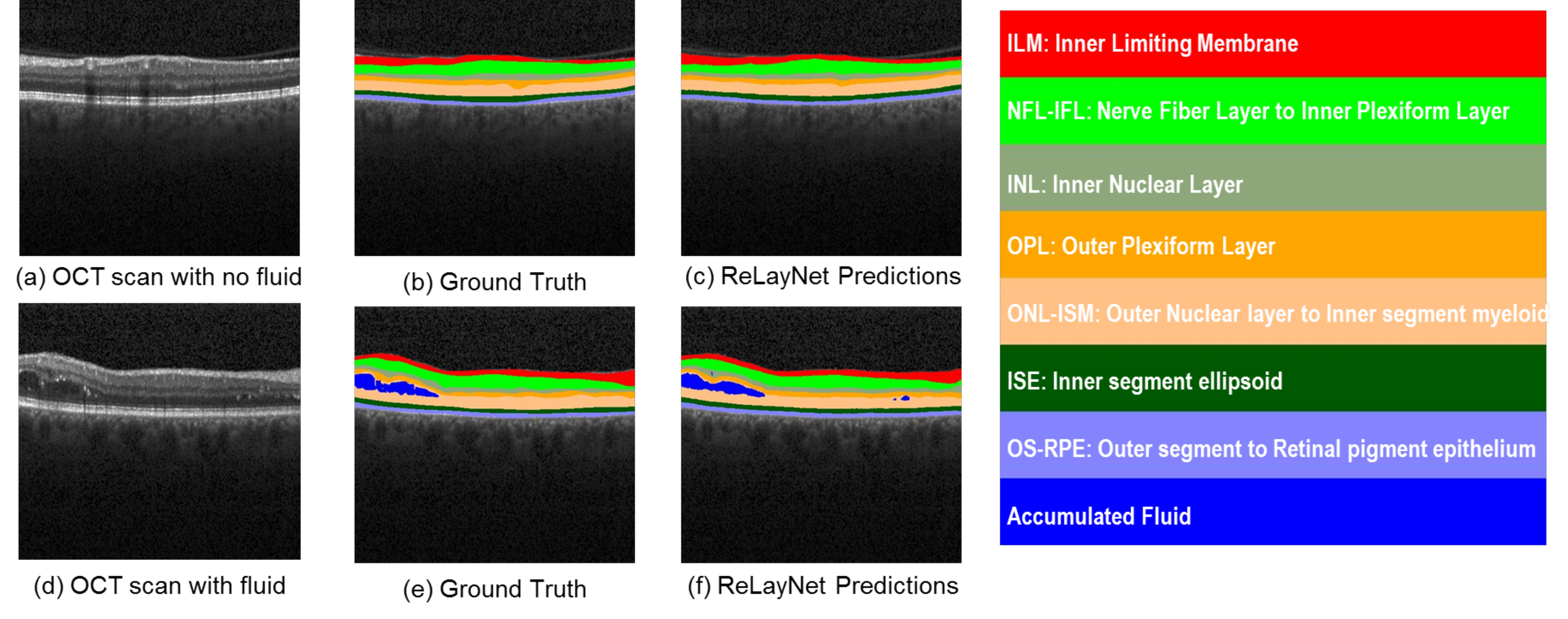}
\caption{Segmentation results of the proposed ReLayNet of OCT frames without and with fluid mass. OCT frame without fluid, its ground truth and ReLayNet segmentation are shown in (a), (b) and (c) respectively. OCT frame with fluid, its ground truth and RelayNet predictions are shown in (d), (e) and (f) respectively. The retinal layers and fluid corresponding to each colors are presented to the right.}
\label{fig:preview}
\end{figure}

Towards this end, we propose a deep learning based end-to-end learning framework for segmentation of multiple retinal layers and delineation of fluid pockets in eye OCT images, called ReLayNet (short for \textit{Re}tinal \textit{Lay}er segmentation network). To the best of our knowledge, this is the first time deep learning based fully convolutional end-to-end method leveraged towards this application. Figure~\ref{fig:preview} previews the results of the proposed ReLayNet for two OCT slices (without and with fluid mass).

\section{State of the art}
\label{sec:stateofart}
The task of segmenting a retinal OCT scan involves partitioning the image into the constituent retinal layers and delineating the fluid pool (if present). OCT is often posed as a graph (called as Graph Construction (GC)) and layer label assignment is solved with dynamic programming (DP) approaches~\cite{kafieh2013,chiu2010,srinivasan2014,chiu2015kernel,dufour2013graph}. Particularly, Chiu \textit{et al.} used intensity gradients to estimate the graph edge-weights followed by DP to solve the shortest path problem, thus estimating the layer boundaries~\cite{chiu2010}. It was further improved in a subsequent method by using hard and soft constraints to add prior information from a learned model within GC~\cite{dufour2013graph}. Alternately, Srinivasan \textit{et al.} proposed using sparsity based image denoising, support vector machines and heuristic priors in GC~\cite{srinivasan2014}. In a relatively recent work, Chiu \textit{et al.} demonstrates the use of kernel regression based methods to classify the layers and fluid masses followed by refinement with GC and DP~\cite{chiu2015kernel}. In similar lines, Karri \textit{et al.} proposed to reinforce GC by learning layer specific edges using structured random forests~\cite{karri2016learning}. Also, spatial consistency across consecutive OCT frames was improved by incorporating appropriate constraints within the DP paradigm for segmentation~\cite{tian2016performance}. Recently, Fang \textit{et al.} combined CNNs with graph search methods for automatically segmenting nine retinal layer boundaries. Their proposed approach uses the probabilistic predictions generated by the CNN within a graph search method that delineates final retinal layer boundaries~\cite{fang2017}.

Parallel approaches inspired by early developments in image segmentation in computer vision applications have also been investigated for OCT segmentation. These include the use of texture information together with diffusion maps~\cite{kafieh2013}, probabilistic approach for modelling retinal layers using layer boundary specific shape regularizers~\cite{rathke2014probabilistic} and deployment of two parallel active contours acting simultaneously to segment the retinal boundaries~\cite{rossant2015parallel}. 

The aforementioned approaches towards retinal layer segmentation are not end-to-end paradigms. Often, heuristics and hand-crafting is employed in choosing the graph weights GC and subsequent DP. The segmentation is achieved in multiple stages involving pre-processing stages like denoising followed by post-processing stages of refinement. Though these additional steps do not limit the usability of these methods, it must be noted that these require significant domain knowledge and modeling approximations. Methods exclusively targeted at layer segmentation often did not consider the presence of fluid filled regions which could lead to potentially erroneous results in pathological settings. In addition to the above limitations, the testing phase of these methods is typically slow (with the graph optimization often being the computational bottleneck). This limits their potential for deployment in time-constrained settings like during interventions. With the main focus of addressing these issues,in this work we propose a deep learning based approach towards generating layer segmentations of a whole B-scan slice in an end-to-end fashion. Deep learning based approaches provide the advantage of learning discriminative representations from the data sans the need for handcrafting features. In particular, we propose a deep learning architecture that falls under the family of fully convolutional neural networks (F-CNN) that are specifically tailored for semantic segmentation which predicts the label for all the image pixel together~\cite{long2015fully,noh2015learning,ronneberger2015u}. 

Of late, there has been considerable amount of work for semantic segmentation using deep learning methods within the computer vision and medical imaging communities. The seminal work on fully convolutional semantic segmentation proposed by Long \textit{et al.}~\cite{long2015fully} is particularly relevant in the context of this work. They effectively adapt state of the art networks trained for image classification into fully convolutional networks fine-tuned for segmentation tasks. Particularly, they introduced the notion of \textit{skip connections} that effectively combines higher-order semantic information from deeper coarsely-resolved layers together with appearance information from shallow finely-resolved layers to improve segmentation detail. A significant improvement over within this family of models was achieved by using an encoder-decoder based framework, termed as DeconvNet and the introduction of unpooling layers instead of interpolation to improve the spatial consistency of segmentation~\cite{noh2015learning}. In an alternate work, Chen \textit{et al.} proposed the concept of using atrous convolutional kernels instead of interpolation to get much smoother version of feature maps that are better suited for semantic segmentation~\cite{chen2016deeplab}. Within the medical imaging community, Ronnerberger \textit{et al.} proposed the U-Net architecture that leverages an encoder-decoder architecture and introduces skip connections across them~\cite{ronneberger2015u}. They demonstrate that such an architecture can be trained effectively in presence of limited training data when appropriate data augmentation and gradient-weighting schemes are employed. It must be noted that the architecture presented in this work is inspired in part by U-Net~\cite{ronneberger2015u} and DeconvNet~\cite{noh2015learning}.

 The salient contributions presented in the paper can be listed as:
(i) To the best of our knowledge, this is the first work employing fully-convolutional deep learning approach for retinal OCT layer and fluid segmentation, (ii) ReLayNet is an end-to-end learning approach that is driven entirely by the OCT data without employing any heuristics or hand-crafting of features and is has highly competitive testing time (\~ 10 ms per B-Scan), (iii) Our model uses an encoder-decoder configuration which is tailored for the task at hand by incorporation of unpooling stages with skip connections for improved spatial consistency and ease of gradient flow during training, (iv) ReLayNet is trained with a composite loss function comprising of a weighted logistic regression loss along with Dice loss for improved segmentation. The weighting scheme employed effectively compensates for imbalanced classes and selectively penalizes misclassification at layer boundaries.

In the rest of the paper, we detail the methodology of the proposed framework in Sec.~\ref{sec:problem}, followed by the experimental setup in Sec.~\ref{sec:expSetup}, results analyzing segmentation performance and retinal thickness estimation are discussed in Sec.~\ref{sec:ObsAndDis} and finally concluding remarks in Sec.~\ref{sec:Conc}.

\section{Methodology}
\label{sec:problem}
\subsection{Problem statement}
Given a retinal OCT image $\mathcal{I}$, the task is to assign each pixel location $\mathbf{x}=(r,c)$ to a particular label $l$ in the label space $\mathcal{L}=\{l\} = \{ 1, \cdots, K \}$ for $K$ classes. We treat the current segmentation task as a $K=10$ class classification problem. The tissue classes include 7 retinal layers illustrated in Figure~\ref{fig:preview}, Region above the retina (RaR), Region below RPE (RbR) and accumulated Fluid.

 \begin{figure}[t]
\centering
\includegraphics[width=\textwidth]{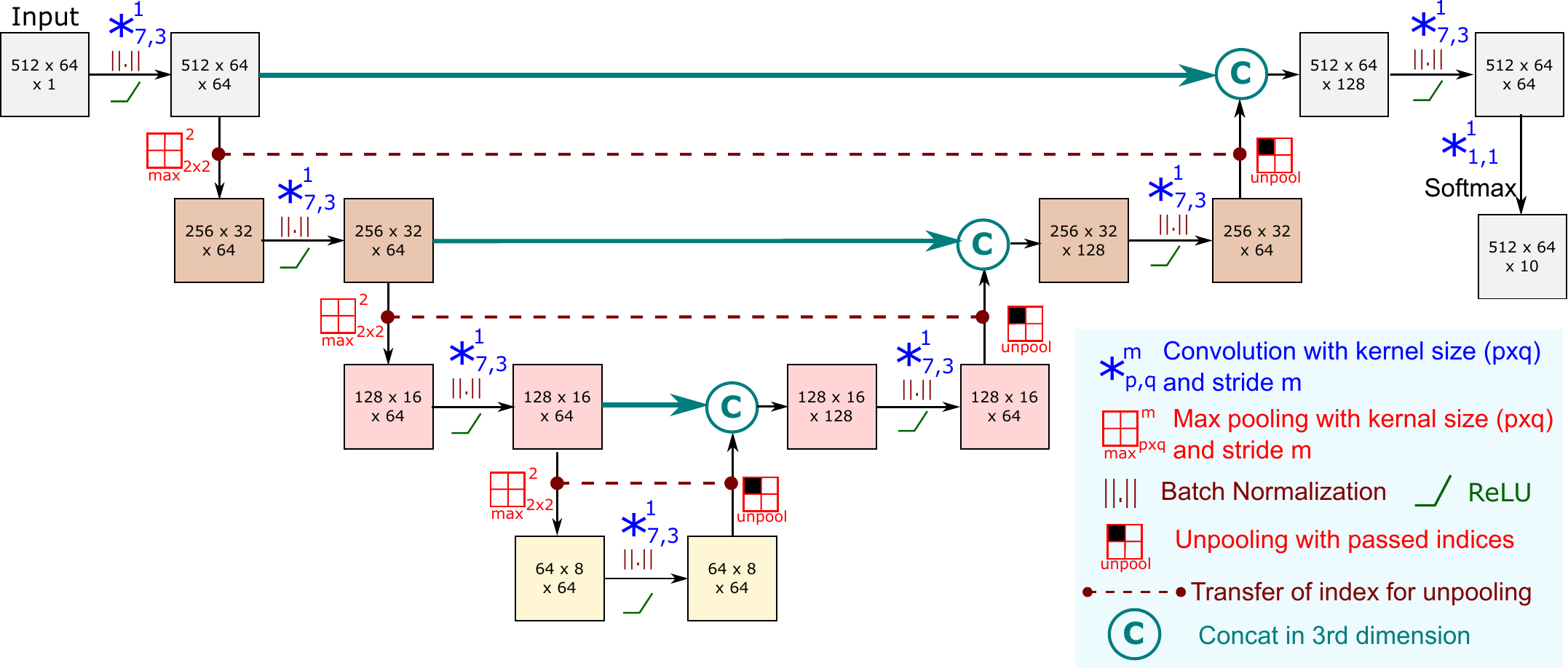}
\caption{Proposed fully convolutional ReLayNet architecture. The spatial resolution of the feature maps are indicated in the boxes. The underlying layer symbols are indicated to the right.}
\label{fig:architecture}
\end{figure}

\subsection{Network architecture}

The network architecture of the proposed ReLayNet is illustrated in Figure~\ref{fig:architecture}. It consists of a contracting path of encoder blocks followed by an expansive path of decoder blocks with skip connections relaying the intermittent feature representations from encoder blocks to their matched decoder blocks through concatenation layers, followed by a classification layer. The individual constituent blocks are detailed as follows:

\subsubsection{Encoder block}
Each encoder block consist of 4 main layers, in sequence: convolution layer, batch normalization layer, ReLU activation layer and max pooling layer. The convolution kernels for all the encoder blocks are defined of rectangular size $7\times3$ to be consistent with OCT image dimensions, along with bias. The kernel size is chosen ensuring that the receptive field at the last encoder block encompasses the whole retinal depth. The feature maps are appropriately zero padded so that the dimension before and after the convolution layer remains the same. A batch normalization layer is introduced after the convolution layer to compensate for the covariate shifts and prevent over-fitting during the training procedure~\cite{ioffe2015batch}. ReLU introduces non-linearity in the training. This is followed by a max pooling layer which reduces the feature map dimensions by half. The pooling indices during this operations are stored and  used in the corresponding unpooling stage of decoder block to preserve spatial consistency.

\begin{figure}[t]
\centering
\includegraphics[height=0.2\textwidth]{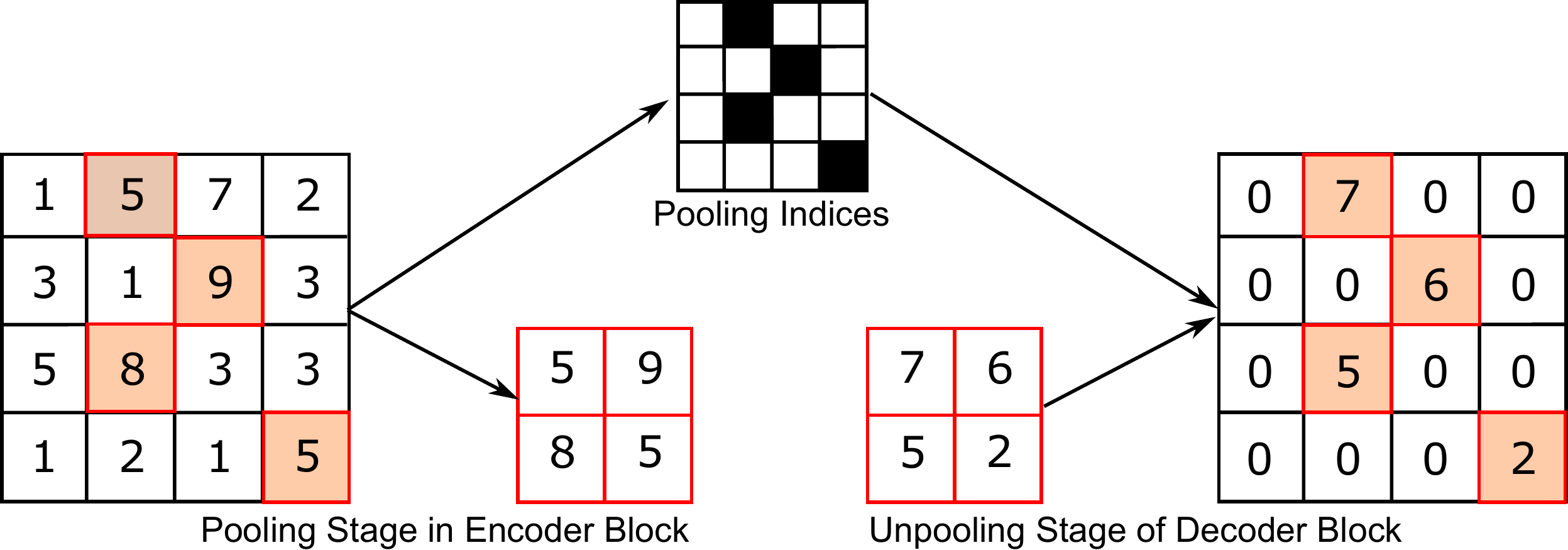}
\caption{Illustration of pooling and unpooling procedures. The pooling stage involves saving the intermediate pooling indices, which is leveraged in the unpooling stage preserving appropriate spatial locations.} 
\label{fig:unpool}
\end{figure}

\subsubsection{Decoder block}
Each decoder block consists of 5 main layers, in sequence: unpooling layer, concatenation layer, convolution layer, batch normalization and ReLU activation function. The unpooling layer upsamples a coarsely-resolved feature map from the preceding decoder block to a finer resolution by using saved pooling indices from the matched encoder block and imputes zeros at the rest of the locations (schematically shown in Figure~\ref{fig:unpool}). Such an unpooling layer ensures that spatial information remains preserved, in contrast to using interpolation for upsampling~\cite{noh2015learning}. This is of particular importance for accurately segmenting layers near the fovea as they are often just a few pixels thick and bilinear interpolation could potentially lead to highly diffused boundaries and hence unreliable estimation of layer thickness. This unpooling layer is followed by a skip connection that relays the output feature maps of the matched encoder block which are in turn concatenated with the unpooled feature maps within the concatenation layer. The advantage of such a skip connection is two fold: (i) it aids the transition to finer resolution by adding an information rich feature map from the encoder part, and (ii) it aids the flow of gradients to the encoder part during training, thus minimizing the risk of vanishing gradient as model depth increases. The concatenated feature map is followed by convolutional layer, batch normalization and ReLU. These layers in effect densify the sparse unpooled feature maps. The kernel size of convolution layer is kept constant at $7\times3$ with appropriate padding similar to the encoder blocks.

\subsubsection{Classification block}
The final decoder block is followed by a convolutional layer with $1\times1$ kernels (used for reducing channels of the feature map without changing spatial dimensions) to map the $64$ channel feature map to a $10$ channel feature map (for 10 classes). At the end, a softmax layer estimates the probability of a pixel belonging to either of the 10 classes.

\subsection{Training}

\subsubsection{Loss functions}
The ReLayNet is trained by jointly optimizing the following loss functions:

\noindent
\textbf{Weighted multi-class logistic loss:} Cross-entropy provides a probabilistic similarity between the actual label and the predicted value at the current state of the network. The average cross-entropy of all the classes defines the logistic loss, which penalizes at each pixel location $\mathbf{x}$ the deviation of the estimated probability $p_{l}(\mathbf{x})$ from 1 and is defined as follows: 
\begin{equation}
\mathcal{J}_{\mathrm{log loss}} = -\sum_{\mathbf{x}\in \Omega} \omega(\mathbf{x})g_{l}(\mathbf{x})  \log(p_{l}(\mathbf{x}))
\label{eq:CE}
\end{equation}
\noindent where $p_{l}(\mathbf{x})$ provides the estimated probability of pixel $\mathbf{x}$ to belong to class $l$, and $\omega(\mathbf{x})$ is the weight associated with pixel $\mathbf{x}$. where $g_{l}(\mathbf{x})$ is a vector with one for the true label and zero entries for the others representing the ground truth probability of pixel at location $\mathbf{x}$ to belong to class $l$.
\noindent
We utilize a weighted version of logistic loss for our application to compensate for class-imbalance and encourage kernels that are discriminative towards layer transitions. 

\noindent
\textbf{Dice loss:} Along with the multi-class logistic loss function, we use Dice loss that evaluates spatial overlap with ground-truth. 

Here, we use a differentiable approximation of dice loss, defined as follows~\cite{milletari2016v}: 
\begin{equation}
\mathcal{J}_{\mathrm{dice}} =  1 - \frac{2 \sum_{\mathbf{x}\in \Omega} p_{l}(\mathbf{x}) g_{l}(\mathbf{x})}{\sum_{\mathbf{x}\in \Omega} p_{l}^2(\mathbf{x}) + \sum_{\mathbf{x}\in \Omega} g_{l}^2(\mathbf{x})}
\end{equation}

\subsubsection{Weighting scheme for loss function}
\label{ss:weighting}

\begin{figure}[t]
\centering
\includegraphics[height=0.3\textwidth]{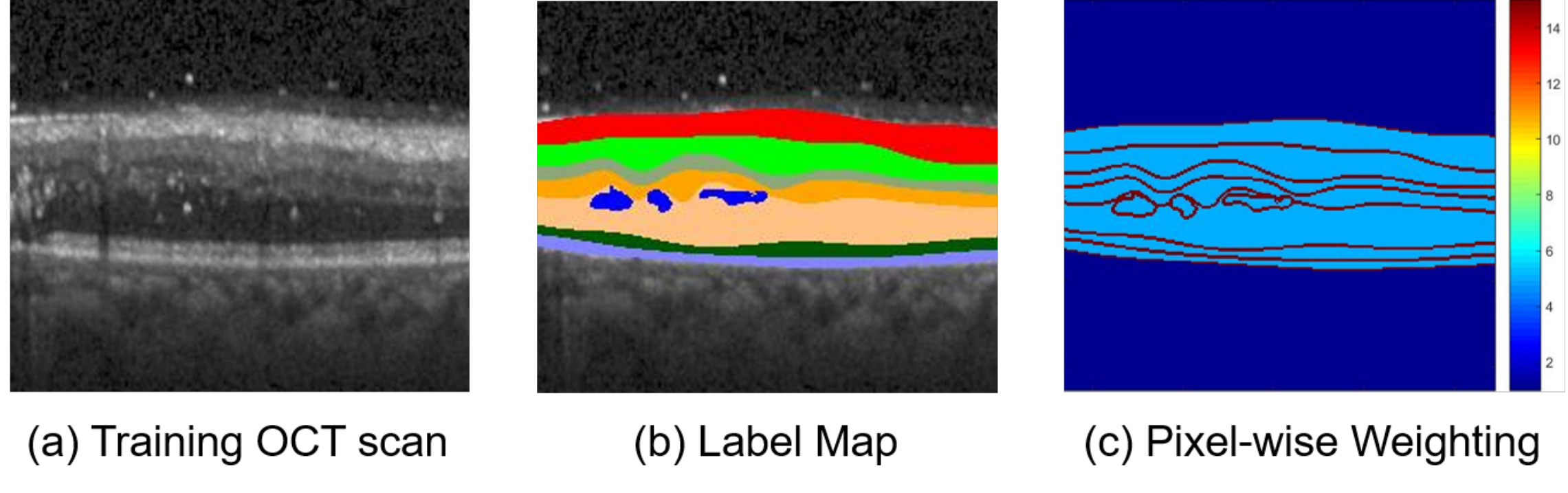}
\caption{Illustration of the weighting scheme for different pixels of a training B-scan OCT image. A sample OCT training B-scan is shown in (a), with its ground truth labels in (b) and the corresponding weights for training as heat map in (c). The color scheme in (b) is consistent with Figure~\ref{fig:preview}.}
\label{fig:weights}
\end{figure}

Let $\omega(\mathbf{x})$ be the weight associated with a particular pixel $\mathbf{x}\in \Omega$ in Eq.~\eqref{eq:CE}. The pixels proximal to tissue-transition regions as evidenced from the ground-truth annotations are often the most challenging to accurately segment as the tissue boundaries may be diffused owing to speckle noise and limited OCT resolution. To encourage the ReLayNet kernels to be sensitive to these, we boost the gradient contributions from such pixels by weighing them with a factor of $\omega_1$. The retinal layers are also heavily imbalanced in contrast to the dominant class (background) and the weighting scheme also aims at compensating this by weighing the contribution of under-represented classes (retinal layers and fluid masses) with a factor of $\omega_2$. Thus, the final weighting scheme is derived as follows (illustrated in Figure~\ref{fig:weights}):   

\begin{equation}
\omega(\mathbf{x}) = 1 + \omega_1 \mathbf{I}(|\nabla l(\mathbf{x})|>0) + \omega_2 \mathbf{I}(l(\mathbf{x})=\mathbf{L}) 
\label{eq:weight}
\end{equation}
\noindent where $\mathbf{I}(logic)$ is a indicator function which is one if the $(logic)$ is true, else returns zero. '$\nabla$' represents the gradient operator. 

\subsection{Optimization}
During training the ReLayNet, we optimize these losses with an additional weight decay term for regularization, defined as follows:
\begin{equation}
\mathcal{J}_{\mathrm{overall}} = \lambda_1 \mathcal{J}_{\mathrm{logloss}} + \lambda_2 \mathcal{J}_{\mathrm{dice}} + \lambda_3 \| \mathbf{W}^{(\cdot)} \|_F^2
\label{eq:lossOverall}
\end{equation}
with weight terms $\lambda_1$, $\lambda_2$ and $\lambda_3$ and $\| \mathbf{W}^{(\cdot)} \|_F$ represents the Frobenius norm on the weights $\mathbf{W}$ of the ReLayNet. The training problem is to estimate the weights and bias $\Theta = \{ \mathbf{W}^{(\cdot)},\mathbf{b}^{(\cdot)} \}$ associated with all the layers, so that to minimize the overall cost function:
\begin{equation}
\Theta^* = \underset{\Theta:\{ \mathbf{W}^{(\cdot)},\mathbf{b}^{(\cdot)} \}}{\mathrm{argmin}} \mathcal{J}_{\mathrm{overall}}(\Theta)
\end{equation}
\noindent
where $\Theta^*$ is the optimal parameter set that minimizes the overall cost. This cost function is optimized using stochastic mini-batch gradient descent with momentum and back propagation. The derivative of the cost function w.r.t. the parameters $\Theta$ is given by:
$\frac{\delta \mathcal{J}_{\mathrm{overall}}}{\delta \Theta} = \frac{\delta \mathcal{J}_{\mathrm{overall}}}{\delta p_{l}(\mathbf{x})}  \frac{\delta p_{l}(\mathbf{x})}{\delta \Theta}$. The second term, $\frac{\delta p_{l}(\mathbf{x})}{\delta \Theta}$ is estimated via chain rule by back propagating the gradients. The first term, $\frac{\delta \mathcal{J}_{\mathrm{overall}}}{\delta p_{l}(\mathbf{x})}$ is estimated as:
\begin{equation}
 \frac{\delta \mathcal{J}_{\mathrm{overall}}}{\delta p_{l}(\mathbf{x})}  = \lambda _{1}\frac{\delta \mathcal{J}_{\mathrm{logloss}}}{\delta p_{l}(\mathbf{x})} + \lambda _{2}\frac{\delta \mathcal{J}_{\mathrm{dice}}}{\delta p_{l}(\mathbf{x})}
\end{equation}
The derivative terms of the individual losses are derived as: 

\begin{equation}
\frac{\delta \mathcal{J}_{\mathrm{logloss}}}{\delta p_{l}(\mathbf{x})} = - \sum_{\mathbf{x}\in \Omega} \frac{\omega(\mathbf{x}) g_{l}(\mathbf{x})}{p_{l}(\mathbf{x})}
\end{equation}
\begin{equation}
\frac{\delta \mathcal{J}_{\mathrm{dice}}}{\delta p_{l}(\mathbf{x})} = -2 \frac{g_{l}(\mathbf{x}) (\sum_{\mathbf{x}\in \Omega} p_{l}^2(\mathbf{x}) + \sum_{\mathbf{x}\in \Omega} g_{l}^2(\mathbf{x})) - 2 p_{l}(\mathbf{x}) (\sum_{\mathbf{x}\in \Omega} p_{l}(\mathbf{x}) g_{l}(\mathbf{x}))}{(\sum_{\mathbf{x}\in \Omega} p_{l}^2(\mathbf{x}) + \sum_{\mathbf{x}\in \Omega} g_{l}^2(\mathbf{x}))^2}
\end{equation}

\begin{figure}[t]
\centering
\includegraphics[width=\textwidth]{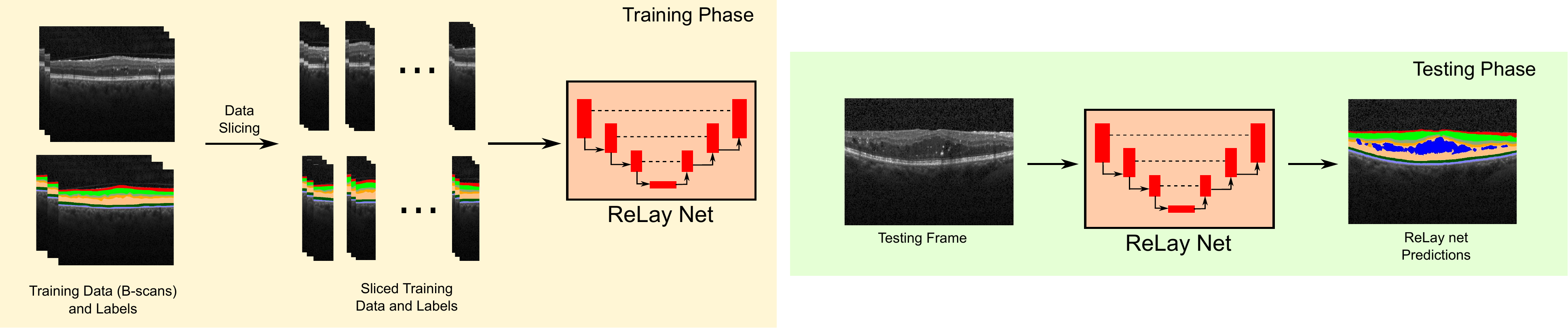}
\caption{Overall flow of the training and testing procedure for the proposed ReLayNet. The training procedure involves slicing of the OCT B-scans as shown above. In testing phase, the whole B-scan is segmented end-to-end.}
\label{fig:GA}
\end{figure}

\subsection{OCT B-scan slicing and data augmentation}
\label{sec:augment}
Training of a deep ReLayNet model with full-width OCT images is limited by the available RAM in the GPU. This requires us to train with smaller batch size, but it often leads to very noisy gradients while training and the loss curve tend to diverge~\cite{ioffe2015batch}. To address this issue, we used a data slicing approach wherein an OCT B-scan is sliced width-wise into a set of non-overlapping B-scan lines as shown in Figure~\ref{fig:GA}. Further, we augment the sliced data by introducing random horizontal flips and slight spatial translations and cropping. It must be noted that due to the resolution-preserving nature of ReLayNet, during test time we use the whole B-scan image, thus obtaining a seamless segmentation without any slicing induced artifacts as shown in Figure~\ref{fig:GA}.

\section{Experimental setup}
\label{sec:expSetup}
\subsection{Dataset}
The proposed framework is evaluated on the Duke SD-OCT publicly available dataset for DME patients~\cite{chiu2015kernel}. The dataset consists of 110 annotated SD-OCT B-scan images of size $512\times740$ obtained from 10 patients suffering from DME (11 B-scans per patient). The 11 B-scans per patient were annotated centered at fovea and 5 frames on either side of the fovea (foveal slice and scans laterally acquired at $\pm 2, \pm 5, \pm10, \pm 15 \text{ and } \pm 20$ from the foveal slice). These 110 B-scans are annotated for the retinal layers and fluid regions by two expert clinicians. The details of the acquisition process is reported in~\cite{chiu2015kernel}.

\subsection{Experimental settings}

Following the standard convention of splitting this dataset as reported in~\cite{chiu2015kernel}, we divide the data with subject 1-5 in the training set and subject 6-10 in the testing set (55 B-scans in each set). The hyper-parameters in loss function in Eq.~\eqref{eq:lossOverall} were set as $\lambda_1 = 1$, $\lambda_2 = 0.5$ and weight decay $\lambda_3 = 0.0001$. For our experiments we empirically set $\omega_1=10$ and $\omega_2=5$. The SGD optimization is performed in mini batches of size = $50$ slices spliced from the training B-Scans (ref. Sec.~\ref{sec:augment}) with augmentation. During start of training the learning rate is set to $0.1$ and is reduced by an order of magnitude after every $30$ epochs. The training was performed with a momentum of 0.9. The training parameters are kept constant for all the deep learning comparative methods and baselines for fair comparison (discussed in Sec.~\ref{sec:cm_bsl}). All the networks were run till convergence. The training was conducted with expert~1 annotations and expert~2 annotation is used exclusively for validation purposes. The experiments were run in a Work station with Intel Xeon CPU, one 12 GB Nvidia Tesla K40 GPU and 64 GB RAM.

\subsection{Comparative methods and baselines}
\label{sec:cm_bsl}
The performance of the proposed ReLayNet is evaluated against state-of-the-art retinal OCT layer segmentation algorithms, specifically Graph based dynamic programming (GDP)~\cite{chiu2010} (CM-GDP), Kernel regression with GDP~\cite{chiu2015kernel} (CM-KR) and Layer specific structured edge learning with GDP~\cite{karri2016learning}(CM-LSE). To contrast ReLayNet with state of the art deep FCN architectures, we include comparisons with U-Net architecture~\cite{ronneberger2015u} (CM-Unet) and FCN architecture proposed by ~\cite{long2015fully} (CM-FCN). Due to limited training data, we reduced the layer depth of CM-Unet in comparison to the original architecture proposed in~\cite{ronneberger2015u}. For a fair comparison, the depth, kernel size, number of channels are kept identical to ReLayNet. This effectively factors out our incremental contributions (unpooling layers with composite loss for training) as the contributing elements for the contrastive results. 
In addition to the above comparative methods, we also present several plausible variations ReLayNet have been set as baselines for comparison, specifically to highlight the importance of each of the proposed contributions. All the baselines are detailed below, with the salient aspects of each baseline detailed in Table~\ref{tab:blConfig}.

\begin{table}[t]\scriptsize
\caption{Salient attributes of the proposed ReLayNet are depth of the architecture, loss function, skip connection and weighting scheme in loss function. The configuration of the baselines are indicated below.}
\begin{center}
  \begin{tabular}{ | p{1cm} | c | c | c | c |}
    \hline
     & Architecture & Loss function & skip connection & Weighting scheme \\ \hline
     BL-1 & $3-1-3$ & Dice + Logistic & $\times$ & \checkmark \\ \hline
     BL-2 & $3-1-3$ & Dice + Logistic & only LR skip & \checkmark \\ \hline
     BL-3 & $3-1-3$ & Dice + Logistic & only HR skip & \checkmark \\ \hline
     BL-4 & $3-1-3$ & Logistic & \checkmark & \checkmark \\ \hline
     BL-5 & $3-1-3$ & Dice & \checkmark & $\times$ \\ \hline
     BL-6 & $2-1-2$ & Dice + Logistic & \checkmark & \checkmark \\ \hline
     BL-7 & $4-1-4$ & Dice + Logistic & \checkmark & \checkmark \\ \hline
     BL-8 & $3-1-3$ & Logistic & \checkmark & $\times$ \\ \hline
  \end{tabular}
\end{center}
\label{tab:blConfig}
\end{table}

\subsection{Evaluation metrics}
\label{sec:evalmet} 
The comparative analysis of segmentation performance is done based on 3 standard metrics as reported in ~\cite{karri2016learning,chiu2015kernel}. These include the Dice overlap score (DS), estimated contour error for each layer (CE) and the error in estimated thickness map (MAD-LT) for each layer. The lateral resolution of the OCT B-scans is between $10.9\mu$m to $11.9\mu$m~\cite{chiu2015kernel}. As the Duke SD-OCT dataset does not report the individual scan resolutions, we resort to reporting our error metrics in pixels as the nearest surrogate.

\begin{figure}[t]
\centering
\includegraphics[width=0.75\textwidth]{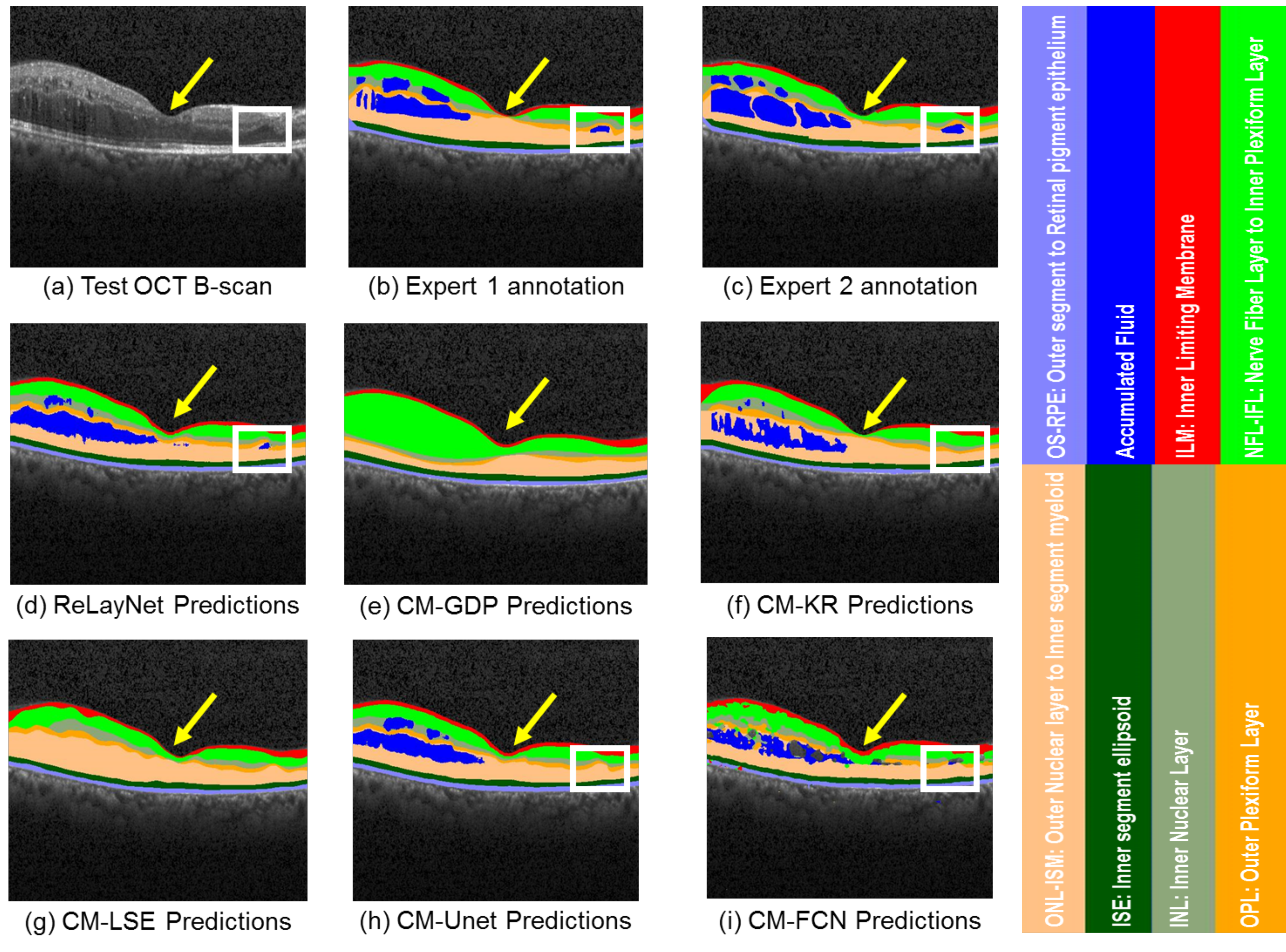}
\caption{Layer and fluid predictions of a Test OCT B-scan near fovea with DME manifestation, shown in (a) with the expert 1 annotations in (b), expert 2 annotations in (c), ReLayNet predictions in (d) and predictions of the defined 5 comparative methods in (e-i). CM-GDP and CM-LSE doesn't include predictions for fluid. The fovea is indicated by the yellow arrow. The region with a small fluid mass is shown by a small white box.}
\label{fig:result}
\end{figure}

\begin{figure}[t]
\centering
\includegraphics[width=0.75\textwidth]{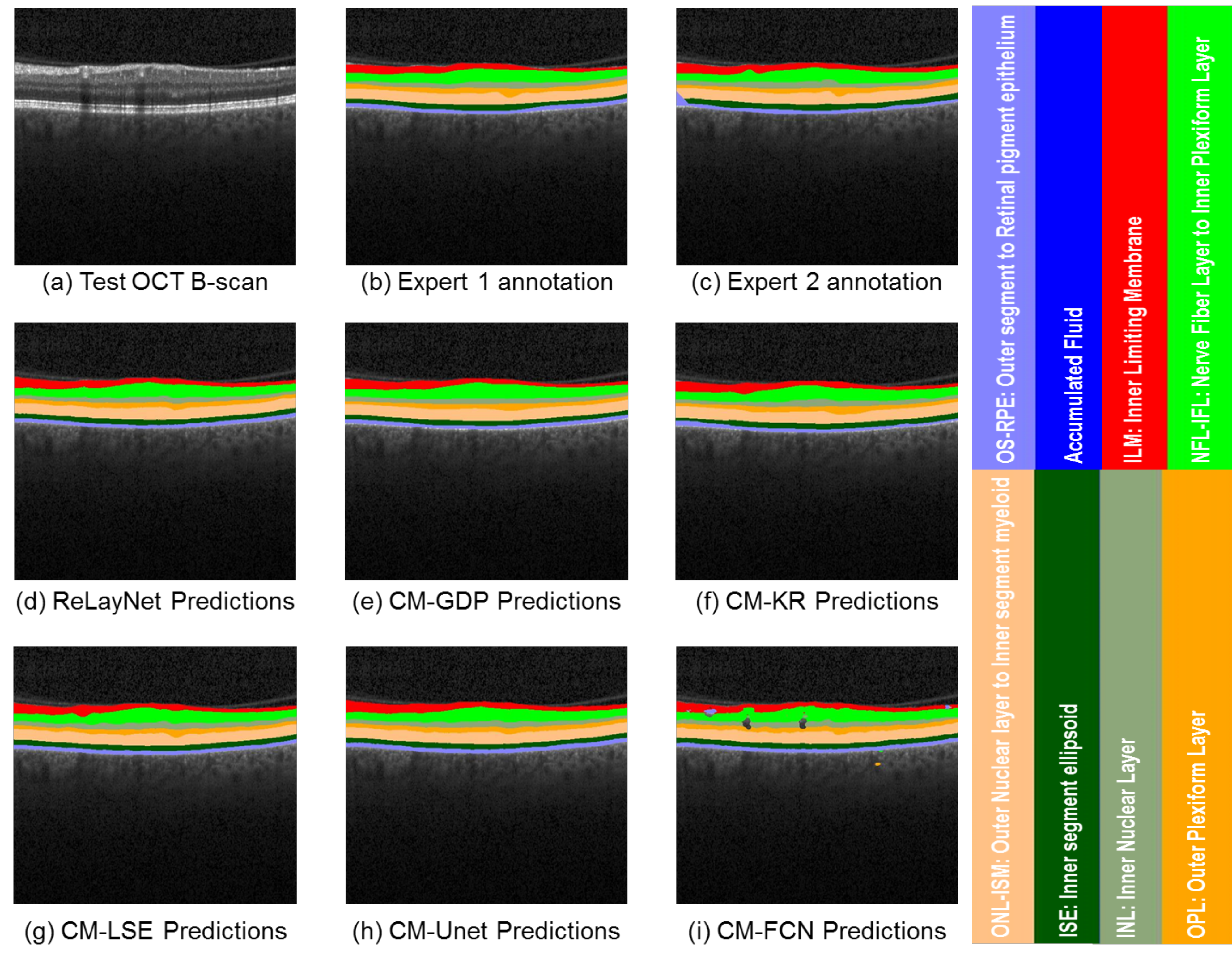}
\caption{Layer predictions of Test OCT B-scan with no fluid mass, shown in (a) with the expert 1 annotations in (b), expert 2 annotations in (c), ReLayNet predictions in (d) and predictions of the defined 5 comparative methods in (e-i).}
\label{fig:result2}
\end{figure}

\section{Experimental observation and discussion}
\label{sec:ObsAndDis}

\subsection{Qualitative comparison of ReLayNet with comparative methods}

We present a qualitative comparison of ReLayNet in contrast with the comparative methods for two cases: an pathological OCT B-scan with DME (as shown in Figure~\ref{fig:result}) and for an OCT  B-Scan sans fluid accumulation distal from the fovea (as shown in Figure~\ref{fig:result2}). 

\noindent
\textbf{OCT B-scan with fluid accumulation: } Foveal scan (presented Figure~\ref{fig:result}(a)) is representative of a challenging pathological case due to the existence of accumulated fluid masses and relatively thin retinal layers at the foveal region (as indicated by yellow arrow in Figure~\ref{fig:result}). We further observe that a small fluid pool towards the right of the B-Scan (shown with a white box) is successfully segmented by ReLayNet, while CM-Unet, CM-FCN and CM-KR prediction fails to capture the small structure (Figure~\ref{fig:result}(b-d)).

Evaluating segmentation performance of layers proximal to the fovea (indicated by yellow arrow), the predictions of CM-GDP is observed to be highly smoothened with lack of detail and it particularly over-predicts Class NFL-IPL and under-predicts the lower retinal layers. In comparison, CM-GDP and CM-LSE result in predictions with greater detail. However, these methods do not consider the presence of fluid while segmenting and the resultant thickness maps may be erroneous at locations proximal to fluid structures. We also observe that the segmentation of ReLayNet and CM-Unet are of high quality and comparable to that of another human expert, indicating the promising potential of F-CNN based frameworks. We also note that CM-FCN performs very poorly at the fluid class and suffers from high confusion between the Fluid and RbR class. This factors the contribution of encoder-decoder based architectures and use of weighted loss function which CM-FCN lacks in comparison to ReLayNet and CM-Unet.

\noindent
\textbf{OCT B-scan without fluid accumulation: } The frame presented in Figure~\ref{fig:result2}(a) is representative of a fluid-free OCT scan acquired distal from the fovea. We can consistent performance across the comparative methods. This observation also indicates that the comparative methods have been trained fairly and the major distinction arises in the presence of pathology where such a objective segmentation tool is often needed. 

\subsection{Quantitative comparison of ReLayNet with comparative methods}
\label{ss:comparative}

Towards quantitative evaluation of performance against the comparative methods, we report the three metrics, namely DS, MAD-LT and CE for each of the retinal layers in Table~\ref{tab:cm-metric}. We also report the DS for the fluid class. From an overall perspective, ReLayNet demonstrates highest segmentation efficacy in 9 classes out of 10 with above 0.9 DS for RaR, ILM, NFL-IPL, ONL-ISM, ISE and OS-RPE layers. CM-Unet has the second best performance for 5 classes among all the comparative methods. In the particular case of ONL-ISM layer, ReLayNet has the second best performance in DS (0.93) which is highly comparable to the best performing comparative method CM-LSE. Further, the OPL layer is the most challenging retinal layer to segment (evident from the low DS of 0.74 between two expert observers). In this challenging layer, ReLayNet achieves a DS of 0.84 which is substantially improved over the other comparative methods with improvements of 0.17, 0.10 and 0.07 over CM-GDP, CM-KR and CM-LSE respectively. In addition to improved layer segmentation, we also observe substantial improvement in the segmentation of fluid masses and report a DS of 0.77. ReLayNet significantly outperforms CM-Unet and CM-FCN in fluid segmentation by margins of 0.10 and 0.49 respectively in DS. CM-FCN exhibits the worst performance of 0.28 DS for fluid class in comparison to all the other comparative methods.

In terms of the MAD-LT metric, ReLayNet achieves consistently superior performance for all the constituent layers. Specifically, CM-GDP has the worst performance in thickness estimation for layers ILM, NFL-IPL, OPL and ONL-ISM. In contrast, CM-KR and CM-LSE exhibit comparable performance and better than CM-GDP due to GC related improvements incorporated in them. Particularly for the ILM layer, the MAD-LT of ReLayNet outperforms CM-GDP, CM-KR and CM-LSE by margins of 2.50, 0.24 and 0.26 pixels respectively. Contrasting with CM-UNet and CM-FCN, we observe improving trends concurrent with the DS metric. With respect to the CE metric, ReLayNet exhibits the best performance for all the layers except ONL-ISM, where CM-LSE and CM-GDP outperforms the proposed method by a margin of 1.23 and 0.97 pixels. This is because CM-GDP and CM-LSE do not involve estimation of the fluid class. The presence of fluid masses, typically within ONL-ISM layer challenges the contour estimation for this layer. It can be observed that ReLayNet outperforms CM-KR by a margin of 0.39 pixels which also involves estimating fluid class. 

Our overall qualitative and quantitative analyses substantiate that ReLayNet performs better than comparative methods based on the introduction of key contributions including that of (i) Dice loss function and (ii) use unpooling layer instead of convolution transpose layers or interpolation which differentiates ReLayNet from CM-Unet and CM-FCN respectively. It also demonstrates that ReLayNet is able to estimate layer thickness better than graph-based comparative approaches and exhibits consistency across pathological variations, despite diffused layer boundaries and presence of speckle noise.

\noindent
\textbf{Comparison with additional human expert observer}:  We also compare the agreeability between the two human expert annotations (Expert 1 vs. Expert 2) and ReLayNet performance (ReLayNet vs. Expert 1) and report the observed metrics in Table~\ref{tab:cm-metric}. The low observer agreement between the two experts reflected particularly by the low DS in retinal layers INL (0.79), OPL (0.74), OS-RPE (0.82) and the fluid class (0.58) shows that the task of retinal segmentation is highly subjective and challenging. This substantiates our premise for the need for an objective solution. Comparing Expert 2 annotations to that predicted by ReLayNet, we can observe a higher agreement with the ground truth (Expert 1) for ReLayNet.

\begin{table}[t]
\scriptsize
\caption{Comparison with Comparative Methods and Expert 2 annotations. The best performance is shown by \textbf{bold}, the second best is shown by $\star$ and the worst shown by $\dagger$.}
\centering
\begin{tabular}{|c|c|p{0.5cm}|p{0.5cm}|c|p{0.5cm}|p{0.5cm}|c|p{0.5cm}|c|p{0.5cm}|p{0.5cm}|}\hline
~ & \multicolumn{1}{c|}{} & \multicolumn{1}{c|}{\textbf{RaR}} & \multicolumn{1}{c|}{\textbf{ILM}} & \multicolumn{1}{c|}{\textbf{NFL-IPL}} & \multicolumn{1}{c|}{\textbf{INL}} & \multicolumn{1}{c|}{\textbf{OPL}} & \multicolumn{1}{c|}{\textbf{ONL-ISM}} & \multicolumn{1}{c|}{\textbf{ISE}} & \multicolumn{1}{c|}{\textbf{OS-RPE}} & \multicolumn{1}{c|}{\textbf{RbR}} & \multicolumn{1}{c|}{\textbf{Fluid}} \\ \hline
\multirow{7}{*}{\rotatebox[origin=c]{90}{\parbox[c]{1cm}{\centering Dice}}} &&&&&&&&&&&\\
    &Proposed & $\mathbf{0.99}$ & $\mathbf{0.90}$ & $\mathbf{0.94}$ & $\mathbf{0.87}$ & $\mathbf{0.84}$ & $0.93^\star$ & $\mathbf{0.92}$ & $\mathbf{0.90}$ & $\mathbf{0.99}$ & $\mathbf{0.77}$  \\ %\hline
    &Expert 2 & NA & $0.86$ & $0.90$ & $0.79$ & $0.74$ & $\mathbf{0.94}$ & $0.86^\dagger$ & $0.82^\dagger$ & NA & $0.58$  \\ %\hline
    &CM-GDP & NA & $0.77^\dagger$ & $0.77^\dagger$ & $0.65^\dagger$ & $0.67^\dagger$ & $0.86^\dagger$ & $0.87$ & $0.82^\dagger$ & NA & NA  \\ %\hline
    &CM-KR & NA & $0.85$ & $0.89$ & $0.75$ & $0.74$ & $0.93$ & $0.87$ & $0.82^\dagger$ & NA & $0.53$  \\ %\hline
    &CM-LSE & NA & $0.87^\star$ & $0.90$ & $0.80$ & $0.77$ & $\mathbf{0.94}$ & $0.88$ & $0.86^\star$ & NA & NA  \\ %\hline
    &CM-Unet & $\mathbf{0.99}$ & $0.86$ & $0.91^\star$ & $0.83^\star$ & $0.81^\star$ & $0.91$ & $0.90^\star$ & $0.83$ & $\mathbf{0.99}$ & $0.67^\star$  \\ 
    &CM-FCN & $0.97^\star$ & $0.81$ & $0.84$ & $0.72$  &  $0.71$ & $0.88$ & $0.89$ & $0.86$ & $0.98^\star$ & $0.28^\dagger$ \\ \hline
    
    \multirow{7}{*}{\rotatebox[origin=c]{90}{\parbox[c]{1.5cm}{\centering MAD-LT}}} &&&&&&&&&&&\\
    &Proposed & - & $\mathbf{1.50}$ & $\mathbf{1.20}$ & $\mathbf{1.00}$ & $\mathbf{1.31}$ & $\mathbf{1.35}$ & $\mathbf{0.62}$ & $\mathbf{0.92}$ & - & -  \\ %\hline
    &Expert 2 & - & $2.01$ & $2.33$ & $2.17$ & $2.29$ & $2.24$ & $1.53^\dagger$ & $1.54$ & - & -  \\ %\hline
    &CM-GDP & - & $4.03^\dagger$ & $4.04^\dagger$ & $1.99$ & $2.63^\dagger$ & $5.50^\dagger$ & $1.32$ & $1.32$ & - & -  \\ %\hline
    &CM-KR & - & $1.74^\star$ & $2.32$ & $2.27^\dagger$ & $2.38$ & $2.64$ & $1.49$ & $1.49$ & - & - \\ %\hline
    &CM-LSE & - & $1.76$ & $2.25$ & $2.19$ & $2.31$ & $2.31$ & $1.26$ & $1.23$ & - & - \\ %\hline
    &CM-Unet & - & $3.37$ & $1.48^\star$ & $1.27^\star$ & $1.33^\star$ & $2.10^\star$ & $0.71^\star$ & $1.81^\dagger$ & - & - \\ 
    &CM-FCN & - & $2.07$ &  $2.44$ &   $2.14$  &  $1.91$  &  $2.81$  &  $1.30$  &  $1.22^\star$ & - & - \\
    \hline
    
    \multirow{7}{*}{\rotatebox[origin=c]{90}{\parbox[c]{1cm}{\centering CE}}} &&&&&&&&&&&\\
    &Proposed & - & $\mathbf{0.85}$ & $\mathbf{1.14}$ & $\mathbf{1.22}$ & $\mathbf{1.35}$ & $2.09$ & $\mathbf{0.81}$ & $\mathbf{0.81}$ & - & -  \\ %\hline
    &Expert 2 & - & $1.14$ & $1.68$ & $1.68$ & $1.72$ & $1.95$ & $1.10^\dagger$ & $1.27$ & - & -  \\ %\hline
    &CM-GDP & - & $1.09$ & $3.96^\dagger$ & $5.94^\dagger$ & $5.31^\dagger$ & $1.12^\star$ & $1.04$ & $1.35^\dagger$ & - & -  \\ %\hline
    &CM-KR & - & $1.32$ & $1.70$ & $2.01$ & $2.16$ & $2.48^\dagger$ & $1.06$ & $1.18$ & - & - \\ %\hline
    &CM-LSE & - & $0.97^\star$ & $1.62$ & $1.70$ & $2.14$ & $\mathbf{0.86}$ & $1.08$ & $0.86^\star$ & - & - \\ %\hline
    &CM-Unet & - & $1.20$ & $1.15^\star$ & $1.26^\star$ & $1.45^\star$ & $2.13$ & $0.86^\star$ & $1.27$ & - & - \\
    &CM-FCN & - & $1.73^\dagger$  &  $2.72$  &  $2.07$  &  $2.07$  &  $2.84$ & $0.94$  &  $0.96$ & - & - \\
    \hline
\end{tabular}
\label{tab:cm-metric}
\end{table}
 
\subsection{Importance of ReLayNet contributions}
\noindent
\textbf{Importance of skip connections BL-1-3}: Contrasting with BL-1 (sans any skip connections), we observe that ReLayNet outperforms in segmentation performance across all the retinal layers and fluid masses. Particularly, a significant improvement of 0.09 in DS is observed in the fluid class. This is also reflected in MAD-LT and CE of the ONL-ISM layer where ReLayNet improves over BL-1 by a margin of 0.6 and 0.08 pixels respectively. This observed improvement is owed to the introduction of skip connections which improves trainability of deep models and provides additional contextual information derived from encoder-features maps for improved segmentation~\cite{drozdzal2016}. To further understand the relative importance of various levels of skip connections within ReLayNet, we contrast with BL-2 (only coarse-resolution skip connections) and BL-3 (only fine-resolution skip connections). In addition to consistently superior performance of ReLayNet, we particularly observe an increase of 0.08 and 0.11 dice score for the fluid class over BL-2 and BL-3 respectively. Specifically for the ONL-ISM layer, a increase in MAD-LT of 1.6 and 1.5 pixels for BL-2 and BL-3 respectively is observed. These observations affirm our premise that skip connections at all levels of resolution are highly contributory and introducing them induces significant improvements in network performance.

\begin{table}[t]
\scriptsize
\caption{Comparison with baselines. The best performance is shown by \textbf{bold}, the second best is shown by $\star$ and the worst shown by $\dagger$.}
\centering
\begin{tabular}{|c|c|p{0.5cm}|p{0.5cm}|c|p{0.5cm}|p{0.5cm}|c|p{0.5cm}|c|p{0.5cm}|p{0.5cm}|}\hline
~ & \multicolumn{1}{c|}{} & \multicolumn{1}{c|}{\textbf{RaR}} & \multicolumn{1}{c|}{\textbf{ILM}} & \multicolumn{1}{c|}{\textbf{NFL-IPL}} & \multicolumn{1}{c|}{\textbf{INL}} & \multicolumn{1}{c|}{\textbf{OPL}} & \multicolumn{1}{c|}{\textbf{ONL-ISM}} & \multicolumn{1}{c|}{\textbf{ISE}} & \multicolumn{1}{c|}{\textbf{OS-RPE}} & \multicolumn{1}{c|}{\textbf{RbR}} & \multicolumn{1}{c|}{\textbf{Fluid}} \\ \hline
\multirow{9}{*}{\rotatebox[origin=c]{90}{\parbox[c]{1cm}{\centering Dice}}} &&&&&&&&&&&\\
    &Proposed & $\mathbf{0.99}$ & $\mathbf{0.90}$ & $\mathbf{0.94}$ & $\mathbf{0.87}$ & $\mathbf{0.84}$ & $\mathbf{0.93}$ & $\mathbf{0.92}$ & $\mathbf{0.90}$ & $\mathbf{0.99}$ & $\mathbf{0.77}$  \\ %\hline
    &BL-1 & $\mathbf{0.99}$ & $0.84^\dagger$ & $0.92$ & $0.83$ & $0.80$ & $0.89^\dagger$ & $0.90$ & $0.83^\dagger$ & $\mathbf{0.99}$ & $0.68$  \\ %\hline
    &BL-2 & $\mathbf{0.99}$ & $0.85$ & $0.91$ & $0.81^\dagger$ & $0.78^\dagger$ & $0.89^\dagger$ & $0.89^\dagger$ & $0.86$ & $\mathbf{0.99}$ & $0.66^\dagger$  \\ %\hline
    &BL-3 & $\mathbf{0.99}$ & $0.88$ & $0.92$ & $0.83$ & $0.81$ & $0.89^\dagger$ & $0.90$ & $0.87$ & $\mathbf{0.99}$ & $0.69$  \\ %\hline
    &BL-4 & $\mathbf{0.99}$ & $0.85$ & $0.92$ & $0.83$ & $0.81$ & $0.91$ & $0.90$ & $0.83^\dagger$ & $\mathbf{0.99}$ & $0.72$  \\ %\hline
    &BL-5 & $\mathbf{0.99}$ & $0.88$ & $0.92$ & $0.83$ & $0.82^\star$ & $0.92^\star$ & $0.90$ & $0.88$ & $\mathbf{0.99}$ & $0.73$  \\ %\hline
    &BL-6 & $0.97^\star$ & $0.86$ & $0.90^\dagger$ & $0.81^\dagger$ & $0.80$ & $0.90$ & $0.90$ & $0.83^\dagger$ & $\mathbf{0.99}$ & $0.68$  \\ %\hline
    &BL-7 & $\mathbf{0.99}$ & $0.89^\star$ & $0.93^\star$ & $0.84^\star$ & $0.82^\star$ & $0.92^\star$ & $0.91^\star$ & $0.89^\star$ & $\mathbf{0.99}$ & $0.76^\star$  \\ %\hline
    &BL-8 & $\mathbf{0.99}$ & $0.86$ & $0.90^\dagger$ & $0.81^\dagger$ & $0.78^\dagger$ & $0.89^\dagger$ & $0.89^\dagger$ & $0.86$ & $\mathbf{0.99}$ & $0.66^\dagger$  \\ \hline
    
    \multirow{9}{*}{\rotatebox[origin=c]{90}{\parbox[c]{1.5cm}{\centering MAD-LT}}} &&&&&&&&&&&\\
    &Proposed & - & $\mathbf{1.50}$ & $\mathbf{1.20}$ & $\mathbf{1.00}$ & $1.31$ & $\mathbf{1.35}$ & $\mathbf{0.62}$ & $0.92$ & - & -  \\ %\hline
    &BL-1 & - & $2.90$ & $1.33$ & $1.29$ & $1.44$ & $1.95$ & $0.73$ & $1.88^\dagger$ & - & - \\ %\hline
    &BL-2 & - & $1.86$ & $1.40$ & $1.35$ & $1.34$ & $2.92^\dagger$ & $0.72$ & $1.00$ & - & - \\ %\hline
    &BL-3 & - & $1.73$ & $1.70^\dagger$ & $1.45$ & $1.45$ & $2.88$ & $0.71$ & $0.96$ & - & - \\ %\hline
    &BL-4 & - & $3.05^\dagger$ & $1.27^\star$ & $1.25$ & $1.32$ & $1.80$ & $0.65$ & $1.61$ & - & - \\ %\hline
    &BL-5 & - & $1.70$ & $1.45$ & $1.34$ & $1.28^\star$ & $3.18$ & $0.66$ & $0.86$ & - & - \\ %\hline
    &BL-6 & - & $1.71$ & $1.47$ & $1.23^\star$ & $\mathbf{1.04}$ & $2.27$ & $0.69$ & $\mathbf{0.83}$ & - & - \\ %\hline
    &BL-7 & - & $1.60^\star$ & $1.58$ & $1.32$ & $1.42$ & $1.52^\star$ & $0.65^\star$ & $0.96$ & - & - \\ %\hline
    &BL-8 & - & $1.88$ & $1.29$ & $1.71^\dagger$ & $1.71^\dagger$ & $2.40$ & $0.79^\dagger$ & $0.84^\star$ & - & - \\ %\hline
    \hline
    
    \multirow{9}{*}{\rotatebox[origin=c]{90}{\parbox[c]{1cm}{\centering CE}}} &&&&&&&&&&&\\
    &Proposed & - & $\mathbf{0.85}$ & $\mathbf{1.14}$ & $\mathbf{1.22}$ & $\mathbf{1.35}$ & $\mathbf{2.09}$ & $\mathbf{0.81}$ & $\mathbf{0.81}$ & - & -  \\ %\hline
    &BL-1 & - & $1.59^\dagger$ & $1.28$ & $1.32$ & $1.40$ & $2.17$ & $0.85$ & $0.89$ & - & - \\ %\hline
    &BL-2 & - & $1.13$ & $1.26$ & $1.34$ & $1.42$ & $2.24$ & $0.83^\star$ & $0.84$ & - & - \\ %\hline
    &BL-3 & - & $1.12$ & $1.25$ & $1.33$ & $1.45$ & $2.21$ & $0.84$ & $0.85$ & - & - \\ %\hline
    &BL-4 & - & $1.46$ & $1.21$ & $1.27$ & $1.37^\star$ & $\mathbf{2.09}$ & $0.83^\star$ & $1.27^\dagger$ & - & - \\ %\hline
    &BL-5 & - & $1.11$ & $1.23$ & $1.32$ & $1.42$ & $2.25$ & $0.83^\star$ & $0.82^\star$ & - & - \\ %\hline
    &BL-6 & - & $1.32$ & $1.56^\dagger$ & $1.42$ & $1.51$ & $2.30$ & $0.83^\star$ & $0.84$ & - & - \\ %\hline
    &BL-7 & - & $0.90^\star$ & $1.17^\star$ & $1.23^\star$ & $1.37^\star$ & $2.11^\star$ & $\mathbf{0.81}$ & $0.82^\star$ & - & - \\ %\hline
    &BL-8 & - & $1.25$ & $1.47$ & $1.46^\dagger$ & $1.56^\dagger$ & $2.37^\dagger$ & $0.88^\dagger$ & $0.89$ & - & - \\ 
    \hline
\end{tabular}
\label{tab:metric}
\end{table}

\noindent
\textbf{Effect of joint loss functions BL 4-5}: We contrast ReLayNet with BL-4 (only weighted logistic loss) and BL-5 (only dice loss) to ablatively test the effect of the joint loss. From Table~\ref{tab:metric}, we observe that the loss functions are complementary in nature and improve segmentation performance. Particularly for ILM and OS-RPE, BL-5 is better than BL-4 by a margin of 0.03 and 0.05 DS. These two layers represent the two boundaries of retina and are very susceptible to be confused with the background classes. A similar observation is made for retinal thickness estimation (increase in error of 1.3 and 1.6 pixels for ILM and OS-RPE respectively) and contour estimation (increase in error of 0.3 and 0.5 pixels for ILM and OS-RPE respectively) comparing BL-4 vs. BL-5. This effect is much more dominant with the joint action of dice loss along with logistic loss as proposed for ReLayNet.

\noindent
\textbf{Effect of depth of network BL 6-7}: The choice of network depth is closely related to model complexity. Ideally, we need a network that is sufficiently deep to learn discriminative hierarchy of task-specific kernels while simultaneously not over-fitting to the training data. 
In light of this empirically accepted design rule, we explored three plausible architectures with varying depths as comparative baselines to the ReLayNet architecture. These architectures are addressed as $x-1-x$, which symbolizes an architecture with $x$ encoder blocks, $1$ low resolution bottleneck convolutional block connecting the encoder and decoder followed by $x$ matched decoder blocks. The architectures explored are $2-1-2$ (BL-6), $3-1-3$ (ReLayNet) and $4-1-4$ (BL-7). The performance of all these architectures are reported in Table~\ref{tab:metric}. We observe that the DS of BL-6 deteriorates in comparison to ReLayNet. Particularly, a decrease in 0.09 dice is observed for fluid class, with similar trends in MAD-LT and CE. Contrasting with BL-7, we can observe for most of the classes, the dice performance is almost same as ReLayNet except for the classes INL, OPL with an decrease of 0.03 and 0.02 dice scores. We can conclude that though this is not a conclusive case of over-fitting, the additional model complexity with increased depth offers limited improvement and the proposed layer configuration ($3-1-3$) is an optimal.

\noindent
\textbf{Importance of pixel-wise weighted loss function BL-8}: In Sec.~\ref{ss:weighting}, we motivated the weighting scheme to encourage learning kernels sensitive to layer transitions and efficiently compensate class balance. To ablative test such a scheme, we introduce BL-8 sans any weighting (\textit{i.e.} $\omega_1, \omega_2$ are set to 0 in Eq.~\eqref{eq:weight}) and tabulate the observed performance metrics in Table~\ref{tab:metric}. Comparing against BL-4 (network with weighted logistic loss), we observe a fall of 0.06 in DS for the fluid class, which is the most under-represented class within the training data. In terms of MAD-LT, BL-8 exhibits the lowest performance for INL, OPL and ISE amongst all the baselines. For CE, BL-8 reports the lowest performance for INL, OPL, ONL-ISM and ISE amongst all the baselines. This fall in performance reaffirms the importance of a weighting scheme within the loss function as motivated earlier, for better segmentation at layer boundaries and compensating for class imbalance.

\subsection{Folded cross validation}

To fully utilize the benchmark dataset, we performed additional \textit{k- fold} cross-validation by splitting the dataset into non-overlapping subsets of 8 patients for training and held out 2 patients for testing. Within the training dataset, 8-folded cross-validation was performed, resulting in eight independently trained ReLayNet models. We also report the ensemble performance of these folded-models on test data and compare it against the model trained with the standard 50 - 50 \% split (discussed earlier in Sec.~\ref{ss:comparative}). The fold-wise and ensemble results evaluated with standard metrics Dice, MAD-LT and CE are tabulated in Table~\ref{tab:crossval}. We also contrast the ensemble results with the results of the standard split (tabulated as 50 -50 Split in Table~\ref{tab:crossval}) and observe a significant increase of 6\% in the Dice metric for the Fluid class and consistent improvements across the retinal layers. These observations support that testing with ensemble of ReLayNet models leads to improved segmentation performance but the testing time is traded-off. 

\begin{table}[t]
\scriptsize
\caption{Results of 8-Fold Cross Validation on 8 Patients and ensemble performance of 8 models on rest two patients.}
\centering
\begin{tabular}{|c|c|p{0.5cm}|p{0.5cm}|c|p{0.5cm}|p{0.5cm}|c|p{0.5cm}|c|p{0.5cm}|p{0.5cm}|}\hline
~ & \multicolumn{1}{c|}{} & \multicolumn{1}{c|}{\textbf{RaR}} & \multicolumn{1}{c|}{\textbf{ILM}} & \multicolumn{1}{c|}{\textbf{NFL-IPL}} & \multicolumn{1}{c|}{\textbf{INL}} & \multicolumn{1}{c|}{\textbf{OPL}} & \multicolumn{1}{c|}{\textbf{ONL-ISM}} & \multicolumn{1}{c|}{\textbf{ISE}} & \multicolumn{1}{c|}{\textbf{OS-RPE}} & \multicolumn{1}{c|}{\textbf{RbR}} & \multicolumn{1}{c|}{\textbf{Fluid}} \\ \hline
\multirow{10}{*}{\rotatebox[origin=c]{90}{\parbox[c]{1cm}{\centering Dice}}} &&&&&&&&&&&\\
    &Fold 1 & 0.99 & 0.92 & 0.94 & 0.87 & 0.85 & 0.93 &	0.92 &	0.90 &	0.99 &	0.75  \\ %\hline
     &Fold 2 & 0.99 &	0.89 &	0.93 &	0.85 &	0.85 &	0.92 &	0.91 &	0.91 &	0.99 &	0.75  \\ %\hline
    &Fold 3 & 0.99	&0.89	&0.95	&0.88	&0.84	&0.93	&0.93	&0.89	&0.99	&0.76  \\ %\hline 
    &Fold 4 &0.99 &	0.88 &	0.94 &	0.89 &	0.83 &	0.94 &	0.92 &	0.92 &	0.99 &	0.83  \\ %\hline
    &Fold 5 & 0.99 &	0.95 &	0.95 &	0.88 &	0.83 &	0.91 &	0.93 &	0.89 &	0.99 &	0.77  \\ %\hline
    &Fold 6 & 0.99 &	0.88 &	0.91 &	0.84 &	0.84 &	0.94 &	0.90 &	0.90 &	0.99 &	0.79  \\ %\hline
    &Fold 7 & 0.99 &	0.89 &	0.94 &	0.88 &	0.87 &	0.94 &	0.93 &	0.89 &	0.99 &	0.75 \\ %\hline
    &Fold 8 & 0.99 &	0.88 &	0.94 &	0.86 &	0.85 &	0.93 &	0.94 &	0.92 &	0.99 &	0.86  \\ \cline{2-12}
    &50 -50 Split & 0.99 &	0.88 &	0.92 &	0.82 &	0.81 &	0.91 &	0.92 &	0.89 &	0.99 &	0.75  \\ \cline{2-12}
    &Ensemble & 0.99 &	0.91 &	0.95 &	0.88 &	0.87 &	0.94 &	0.94 &	0.92 &	0.99 &	0.81  \\ \hline

    \multirow{9}{*}{\rotatebox[origin=c]{90}{\parbox[c]{1cm}{\centering MAD-LT}}} &&&&&&&&&&&\\
   &Fold 1 & - & 1.40 & 1.43 & 1.05 & 1.37 & 1.44 & 0.53 & 0.52 & - & -  \\ %\hline
   &Fold 2 & - & 1.65 &	1.21 &	1.13 &	1.20 &	1.25 &	0.76 &	0.63
 & - & - \\ %\hline
   &Fold 3 & - & 1.44 &	1.19 &	1.09 &	1.28 &	1.30 &	0.55 &	1.02
 & - & - \\ %\hline
   &Fold 4 & - & 1.65 &	1.09 &	0.94 &	1.14 &	1.21 &	0.66 &	0.83 & - & - \\ %\hline
   &Fold 5 & - & 1.58 &	1.28 &	0.94 &	1.11 &	1.31 &	0.53 &	1.05 
 & - & - \\ %\hline
   &Fold 6 & - & 1.00 &	1.26 &	1.04 &	1.18 &	1.47 &	0.59 &	0.73 & - & - \\ %\hline
   &Fold 7 & - & 0.94 &	0.80 &	1.05 &	0.92 &	1.04 &	0.62 &	0.94
 & - & - \\ %\hline
   &Fold 8 & - & 1.32 &	1.05 &	1.15 &	1.15 &	1.13 &	0.72 &	1.07 & - & - \\ \cline{2-12}
   & 50 -50 Split & - &	1.11 &	1.20 &	1.01 &	1.33 &	1.50 &	0.68 &	0.96 &	- &	-  \\ \cline{2-12}
    &Ensemble & - & 1.04 &	1.07 &	0.93 &	1.17 &	1.12 &	0.63 &	0.85 & - & - \\ \hline
   
    \multirow{9}{*}{\rotatebox[origin=c]{90}{\parbox[c]{1cm}{\centering CE}}} &&&&&&&&&&&\\
    &Fold 1 & - & 1.05 &	1.12 &	1.25 &	1.25 &	1.99 &	0.92 &	0.82
 & - & -  \\ %\hline
    &Fold 2 & - & 0.71 & 1.31 &	1.11 &	1.28 &	1.98 &	0.83 &	0.76
 & - & - \\ %\hline
    &Fold 3 & - & 0.83 &	1.06 &	1.16 &	1.16 &	1.41 &	0.85 &	0.94 & - & - \\ %\hline
    &Fold 4 & - & 0.99 &	1.12 &	0.84 &	1.39 &	2.21 &	1.00 &	0.90 & - & - \\ %\hline
     &Fold 5 & - & 1.05 &	1.17 &	1.17 &	1.34 &	1.19 &	1.20 &	0.85 & - & - \\ %\hline
     &Fold 6 & - & 1.04 &	1.07 &	1.02 &	1.45 &	1.03 &	0.87 &	0.79 & - & - \\ %\hline
      &Fold 7 & - & 0.76 &	1.02 &	1.07 &	1.16 &	1.18 &	0.86 &	0.73 & - & - \\ %\hline
      &Fold 8 & - & 0.87 &	1.18 &	1.15 &	1.35 &	1.94 &	0.94 &	0.82 & - & - \\ \cline{2-12}
      &50 -50 Split & - &	1.01 &	1.16 &	1.32 &	1.45 &	1.94 &	0.74 &	0.78 &	- &	-  \\ \cline{2-12}
      &Ensemble & - & 0.75 &	1.09 &	1.07 &	1.25 &	1.39 &	0.77 &	0.73 & - & - \\ \hline

\end{tabular}
\label{tab:crossval}
\end{table}

\subsection{Analysis of ETDRS grid across patients}

\begin{figure}[t]
\centering
\includegraphics[height=0.25\textwidth]{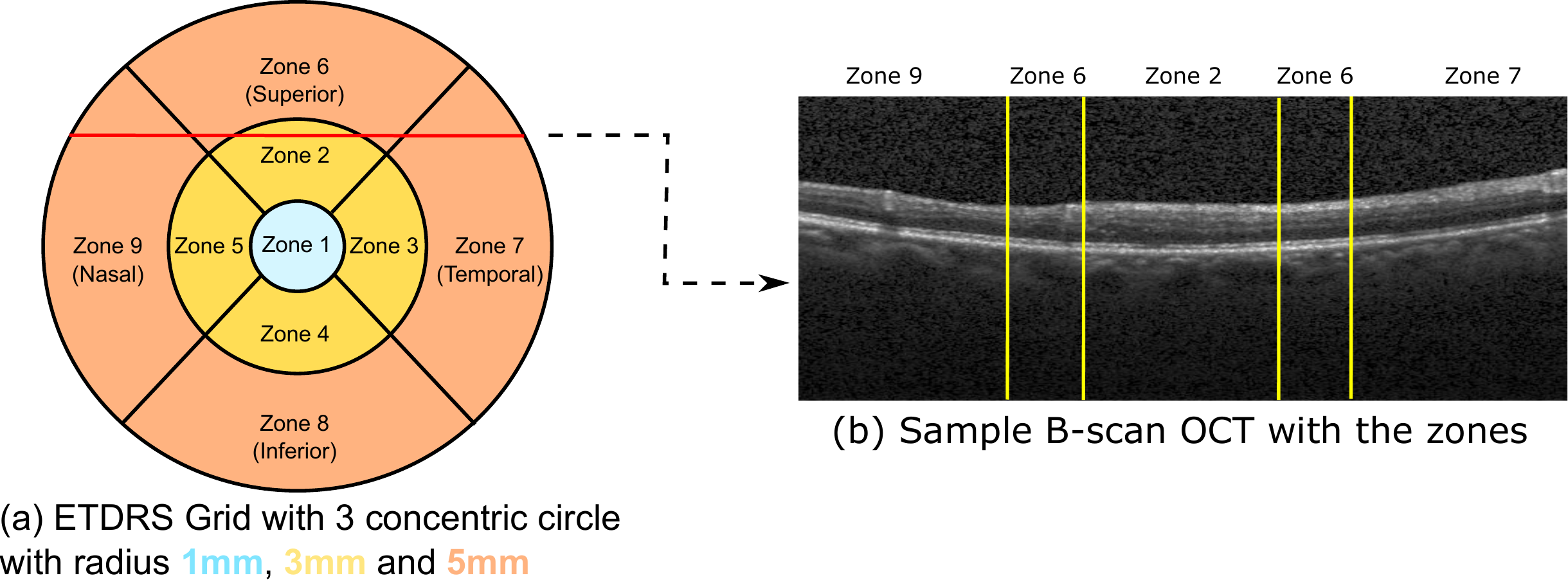}
\caption{Illustration of ETDRS grid with 9 zones as demarcated in (a). This represents the top view for a retinal OCT volume scan. A sample cross-sectional OCT B-scan slice corresponding to the red line in the ETDRS grid is shown in (b). The different regions of the B-scan corresponding to the different zones are indicating by yellow lines in (b).}
\label{fig:etdrs}
\end{figure}

Given a retinal OCT volume acquired for a specific subject, the thickness of the overall retina is generally reported as an Early Treatment of Diabetic Retinopathy Study (ETDRS) grid, providing the average thickness in 9 different spatial zones of the retina~\cite{panozzo2004diabetic}. The spatial zones are centered at the fovea and are delineated as shown in Figure~\ref{fig:etdrs}. The severity of an edema is assessed based on the thickness of these spatial zones by assessing deviations from normal ranges. As the Duke SD-OCT dataset does not contain the local anatomical information (left / right eye) and annotations are made by sparsely sampling the OCT along the azimuthal direction, we assume all the image volumes are aligned left to right with the temporal to nasal axis in our evaluation. The azimuthal resolution of the frames varies between $118$-$128 \mu$m. We assume it as $122 \mu$m for placing it in the grid. It must be noted that due to this assumption, the computed chart is not exact but does show its potential use for computing the grid provided exact resolutions are known. We compute the overall retinal thickness by combining the delineated retinal layers and report the mean absolute difference in overall retinal thickness for the different comparative methods at each of the spatial zones in Table~\ref{tab:etdrs}. We observe that the performance of ReLayNet is significantly better than all the comparative methods for all the 9 zones. Zone 1 indicating the foveal region (most clinically significant) has an very low error of 0.3 pixels. CM-Unet has the second best performance for 4 out of 9 zones (zones 5, 7, 8, 9), especially in regions far from the fovea. The consistent performance of ReLayNet in thickness estimation makes it a good tool for estimating the ETDRS grid.

\begin{table}[t]\scriptsize
\caption{Difference in retinal overall thickness (in pixels) for 9 zones in ETDRS grid across testing subjects. The best performance is shown by \textbf{bold}, the second best is shown by $\star$ and the worst shown by $\dagger$.}
\begin{center}
  \begin{tabular}{ | p{1.5cm} | p{0.8cm} | p{0.8cm} | p{0.8cm} | p{0.8cm} | p{0.8cm} | p{0.8cm} | p{0.8cm} | p{0.8cm} | p{0.8cm} |}
    \hline
     & Zone 1 & Zone 2 & Zone 3 & Zone 4 & Zone 5 & Zone 6 & Zone 7 & Zone 8 & Zone 9  \\ \hline 
    Proposed & $\mathbf{0.34}$ & $\mathbf{0.202}$ & $\mathbf{0.161}$ & $\mathbf{0.204}$ & $\mathbf{0.151}$ & $\mathbf{0.127}$ & $\mathbf{0.123}$ & $\mathbf{0.132}$ & $\mathbf{0.160}$  \\ %\hline
    CM-GDP & $0.59$ & $0.273$ & $0.818^\dagger$ & $0.375$ & $0.698$ & $0.889^\dagger$ & $1.726^\dagger$ & $0.849$ & $2.565$  \\ %\hline
    CM-KR & $1.77^\dagger$ & $0.231$ & $0.474$ & $0.986^\dagger$ & $1.252^\dagger$ & $0.820$ & $1.050$ & $0.965^\dagger$ & $2.576^\dagger$  \\ %\hline
    CM-LSE & $0.64$ & $0.230^\star$ & $0.798$ & $0.289^\star$ & $0.803$ & $0.775$ & $1.189$ & $0.471$ & $2.036$  \\ %\hline
    CM-Unet & $0.89$ & $0.468^\dagger$ & $0.434$ & $0.500$ & $0.368^\star$ & $0.325$ & $0.340^\star$ & $0.327^\star$ & $0.378^\star$  \\ %\hline
    CM-FCN & $0.54^\star$ &   $0.405$ &    $0.398^\star$ &   $0.749$  &  $0.441$ &   $0.214^\star$  &  $0.443$ &   $0.797$  &  $0.461$ \\ \hline
  \end{tabular}
\end{center}
\label{tab:etdrs}
\end{table}

\section{Conclusion}
\label{sec:Conc}

In this paper, we have proposed ReLayNet, an end-to-end fully convolutional framework for semantic segmentation of retinal OCT B-scan into 7 retinal layers and fluid masses. We train and validate it on a publicly available benchmark of expert annotated OCT B-scans acquired from 10 patients. The training of ReLayNet involves minimization of a combined loss comprising of weighted logistic loss and dice loss. ReLayNet is particularly suited for clinical applications owing to its improved test time in the order of 0.01 seconds to segment a single B-Scan.

The proposed ReLayNet framework has been compared and validated against five state-of-the-art retinal layer segmentation methods including ones using graph-based dynamic programming~\cite{chiu2010,chiu2015kernel,karri2016learning} and deep learning~\cite{ronneberger2015u,long2015fully}. Additionally, comparisons have been reported against eight incremental baselines validating each of the individual contributions. The evaluation was performed on the basis of three standard metrics including dice loss, retinal thickness estimation and deviation from layer contours. We demonstrate conclusively that ReLayNet exhibits superior performance in these comparisons and affirm that it can reliably segment even in the presence of a high degree of pathology which severely affects the normal layered structure of the retina. Open questions for future investigation include extension of ReLayNet into intra-operative scenarios like retinal microsurgeries, which poses challenges of poor spatial resolution and artifacts induced by surgical tools. With increasing training data, one could potentially introduce 3D convolutional kernels to improve inter-frame consistency in volume segmentation. 

\section*{Acknowledgements}
This work was supported in part by the Faculty of Medicine at LMU (F\"{o}FoLe), the Bavarian State Ministry of Education, Science and the Arts in the framework of the Centre Digitisation.Bavaria (ZD.B) and we thank the NVIDIA corporation for GPU donation.

\end{document}